\pdfoutput=1

\documentclass[11pt]{article}

\usepackage{acl}

\usepackage{times}
\usepackage{soul}
\usepackage{latexsym}
\usepackage{multirow}
\usepackage{colortbl}
\usepackage{booktabs}
\usepackage{amssymb}
\usepackage{amsmath}
\usepackage{amsthm}
\usepackage{graphicx}

\usepackage{enumitem}
\usepackage[T1]{fontenc}

\usepackage[utf8]{inputenc}

\usepackage{microtype}

\title{Towards Understanding Omission in Dialogue Summarization}

\author{{Yicheng Zou$^{1}$\thanks{{ }{ }This work was done when the first author was an intern at Microsoft Research Asia.}\ ,\ \ Kaitao Song$^2$\thanks{{ }{ }Corresponding authors.}\ ,\ \ Xu Tan$^2$,\ \  Zhongkai Fu$^2$,}\\
{\bf Qi Zhang}$^1$,\ \ {\bf Dongsheng Li}$^2$,\ \ {\bf Tao Gui}$^3$\footnotemark[2]\\
  {$^1$School of Computer Science, Fudan University, Shanghai, China}\\
  {$^2$Microsoft Research Asia, China}\\
  {$^3$Institute of Modern Languages and Linguistics, Fudan University, Shanghai, China}\\
  \texttt{\{yczou18,qz,tgui\}@fudan.edu.cn} \\
  \texttt{\{kaitaosong,xuta,zhongfu,dongsheng.li\}@microsoft.com}}

\newcommand{\song}[1]{{\color{red} #1}}

\begin{document}
\maketitle

\begin{abstract}
Dialogue summarization aims to condense the lengthy dialogue into a concise summary, and has recently achieved significant progress. However, the result of existing methods is still far from satisfactory. Previous works indicated that omission is a major factor in affecting the quality of summarization, but few of them have further explored the omission problem, such as how omission affects summarization results and how to detect omission, which is critical for reducing omission and improving summarization quality. Moreover, analyzing and detecting omission relies on summarization datasets with omission labels (i.e., which dialogue utterances are omitted in the summarization), which are not available in the current literature. In this paper, we propose the \textsc{Olds} dataset, which provides high-quality \underline{O}mission \underline{L}abels for \underline{D}ialogue \underline{S}ummarization. By analyzing this dataset, we find that a large improvement in summarization quality can be achieved by providing ground-truth omission labels for the summarization model to recover omission information, which demonstrates the importance of omission detection for omission mitigation in dialogue summarization. Therefore, we formulate an omission detection task and demonstrate our proposed dataset can support the training and evaluation of this task well. We also call for research action on omission detection based on our proposed datasets. Our dataset and codes are publicly available~\footnote{\url{https://github.com/microsoft/MSummarizer/}}.

\end{abstract}

\section{Introduction}
With the exponential increase in the volume of conversational messages from daily life, there is a growing demand for dialogue summarization~\cite{murray2008summarizing,gliwa2019samsum,chen2020multi,zhong2021qmsum,zou2021low}, which compresses lengthy interactions into a more concise and structured piece of text while preserving the most important and relevant information. Recent years have witnessed significant progress in abstractive dialogue summarization, especially using large-scale pre-trained language models~\cite{lewis2020bart,raffel2020exploring}. Despite the advances in a high level of fluency and coherence, existing models are still prone to generate defective summaries~\cite{kryscinski2019neural,maynez2020faithfulness,tang-etal-2022-confit} that limit their practical usage. Previous works~\cite{chen2020multi,liu2021coreference,tang-etal-2022-confit} have investigated the taxonomy of errors involved in output summaries, and human evaluations revealed that the majority of errors fall into the category of omission, which often leads to incomplete summaries where critical facts are lost. However, few of these works have further analyzed the omission problem, let alone addressing this problem. 

To reduce omission rate and improve summarization quality, a comprehensive analysis on omission problem (e.g., how omission affects summary results) and a precise omission detection (i.e., to locate which dialogue utterances are omitted in the summarization) is important. However, there are no omission related datasets in dialogue summarization literature to support such analysis and detection. Hence, in this work, we construct the \textsc{Olds} dataset, which provides high-quality \underline{O}mission \underline{L}abels for \underline{D}ialogue \underline{S}ummarization. Our dataset is built upon five existing benchmarks covering different domains. For each dialogue, we use different abstractive models to generate diverse candidates and propose a reference-based strategy to automatically label omissions for these candidates. The human evaluation indicates that our \textsc{Olds} dataset presents a high quality of omission labels.

Based on the curated \textsc{Olds} dataset, we comprehensively investigate the omission problem in dialogue summarization from multiple aspects. {\bf First}, we analyze the proportion of candidates with omission errors and the position distribution of omitted information in dialogues. The results reveal that omission is a severe problem that frequently occurs in dialogue summarization. {\bf Second}, we measure the correlation between the omission rate and multiple reference-based metrics (e.g., ROUGE and BERTScore), discovering that omission is one of the decisive factors influencing the summary evaluation results. {\bf Third}, we explore the potential performance improvement brought by utilizing the omission information in a post-editing manner. The analyses probe that candidate summaries could be effectively improved as long as the model is provided with the omitted dialogue utterances. Hence, how to accurately locate omission information in dialogue naturally becomes a critical question. 


To pave the way to omission mitigation and summary improvement, we formulate the task of omission detection, which aims to identify the omitted utterance given the whole dialogue utterances and the generated summary with potential omission. In addition, we present three different frameworks as baselines for the omission detection task, including pair-wise classification, sequence labeling, and pointer network extraction. Experimental analyses on the \textsc{Olds} dataset reveal that omission detection, as a promising direction to assessment and improvement for dialogue summarization, poses significant values and challenges.

The contributions of our paper are as follows:
\begin{itemize}[leftmargin=*]
    \item We propose \textsc{Olds}, a dataset with high-quality omission labels for dialogue summarization, to facilitate the research on the omission problem. 
    \item Based on \textsc{Olds}, we systematically analyze the omission problem and demonstrate the significance of omission in dialogue summarization.
    \item We introduce the omission detection task that paves the way to omission mitigation and summary improvement. We design 3 frameworks as baselines and conduct comprehensive analyses to provide possible directions for solving this task.
\end{itemize}

\begin{table}[t!]
\setuldepth{Berlin}
\fontsize{7.7pt}{9.2pt}\selectfont
\begin{center}
\setlength{\tabcolsep}{0.6mm}{
\begin{tabular}{l}
\toprule[1pt]
\bf \fontsize{8.5pt}{9pt}\selectfont Dialogue: \\
(01) {\bf Adam:} Have you talked to \ul{May}?\\
(02) {\bf Karen:} Yes, yesterday, why?\\
{\bf \color{red}(03)} {\bf \color{red}\ul{Adam}}{\bf :} I just talked to her and I must admit I {\bf \color{red}\ul{worry}} about her.\\
(04) {\bf \ul{Karen}:} Me too, I suggested she should see a specialist, but she \\
\quad \quad wasn't very happy about it.\\
(05) {\bf Adam:} No wonder...\\
(06) {\bf Karen:} I know, but I think this is serious. She's saying she's \\
\quad \quad \ul{depressed}, like everyone around, but in her case it may be true.\\
(07) {\bf Adam:} She was telling me she doesn't feel like doing anything, \\
\quad \quad she's bored all the time, she never feels happy. It sounds like a \\
\quad \quad real, typical depression.\\
\quad \quad {\bf ...... ......}\\
(12) {\bf Adam:} Yes, but she doesn't want to see a specialist. Basically, \\
\quad \quad she doesn't want to see anyone.\\
(13) {\bf \ul{Karen}:} Hm... I don't know... How about I \ul{call} someone for \\
\quad \quad \ul{advice}? So we could know what to do.\\
(14) {\bf Adam:} Sounds rational, do you know anyone you could call? \\
\quad \quad Don't mention her name.\\
{\bf \color{red}(15)} {\bf \ul{Karen}:} Of course I won't! I have a \ul{friend} who's a {\bf \color{red}\ul{psychologist}}, \\
\quad \quad we can trust her. I'll let you know.\\
(16) {\bf Adam:} Thank you Karen!\\
\midrule[0.3pt]
\bf \fontsize{8.5pt}{9pt}\selectfont Reference summary: \\
\quad {\bf \color{red}\ul{Adam}} and \ul{Karen} are {\bf \color{red}\ul{worried}} that \ul{May} suffers from \ul{depression}.  \\
\ul{Karen} will \ul{call} her \ul{friend} who is a {\bf \color{red}\ul{psychologist}} and ask for \ul{advice}. \\
\midrule[0.3pt]
\bf \fontsize{8.5pt}{9pt}\selectfont Candidate summary: \\
\quad \ul{May} is \ul{depressed}. \ul{Karen} suggested she should see a specialist, \\
but she doesn't want to. \ul{Karen} will \ul{call} her \ul{friend} for \ul{advice}. \\
\midrule[0.3pt]
\bf \fontsize{8.5pt}{9pt}\selectfont Omission utterances (Labels): \\
\quad (03) (15)\\
\bottomrule[1pt]
\end{tabular}}
\end{center}
\vspace{-6pt}
\caption{\label{tb:example} An example of the \textsc{Olds} dataset. The dialogue is from SAMSum and the candidate summary is generated from BART$_\mathrm{large}$. The salient words are underlined, and the omission information is highlighted in red. }
\end{table}

\section{The \textsc{Olds} Dataset}
\label{sec:dataset}
In this section, we first define what is omission. Then, we introduce \textsc{Olds}, a dataset that contains \underline{O}mission \underline{L}abels for \underline{D}ialogue \underline{S}ummarization that facilitates the analysis of omission problem and the exploration of  how to identify omission content. Finally, we conduct human assessment that demonstrates the high quality of \textsc{Olds}. 


\begin{table*}[t]
\fontsize{7.8pt}{8pt}\selectfont
\begin{center}
\setlength{\tabcolsep}{0.7mm}{
\begin{tabular}{clcccc|cccccccccccc}
\toprule[1pt]
 \multirow{2}{*}{{\bf Domain}}&\multirow{2}{*}{{\bf Split}}&\bf \# of&\bf Avg.&\bf Len. of& \bf Len. of & \multicolumn{12}{|c}{\bf \# of candidate summaries (Avg. ROUGE-1 score)}\\
 & &\bf dialogs &\bf turns & \bf dialogs & \bf summ. &\multicolumn{2}{|c}{BART$_\mathrm{L}$} & \multicolumn{2}{c}{BART$_\mathrm{B}$} & \multicolumn{2}{c}{T5$_\mathrm{B}$} & \multicolumn{2}{c}{T5$_\mathrm{S}$} & \multicolumn{2}{c}{Transformer} & \multicolumn{2}{c}{Pegasus$_\mathrm{L}$}\\
\midrule
\multirow{3}{*}{{SAMSum}} &Train & 14,732 & 11.2 & 124.1 & 23.4& 25,424 &(50.6)&12,687 &(48.2)&29,473 &(47.2)&32,959 &(41.0)&46,777 &(37.7)&0 &(-)\\
&Dev. & 818 & 10.8 & 121.2 & 23.4 & 1,636 & (54.4)&1,636 &(51.1)&1,636 &(51.0)&1,636 &(44.2)&1,636 &(39.2)&1,636 &(51.1) \\
&Test & 819 & 11.3 & 126.6 & 23.1 & 1,638 &(52.6)&1,638& (49.2)&1,638 &(49.3)&1,638& (43.5)&1,638& (37.9)&1,638 &(50.4) \\
\midrule
\multirow{3}{*}{{DialogSum}} &Train & 12,460 & 9.5 & 187.5 & 31.0& 20,766 &(46.1)&10,132& (44.1)&26,897 &(44.7)&37,056 &(39.5)&29,749& (39.1)&0& (-)\\
&Dev. & 500 & 9.4 & 185.0 & 29.0 & 1,000 &(49.5)&1,000& (46.8)&1,000& (46.2)&1,000& (40.3)&1,000& (40.1)&1,000& (48.4) \\
&Test & 500 & 9.7 & 192.5 & 28.4 & 1,000& (46.9)&1,000& (44.3)&1,000& (44.7)&1,000& (39.1)&1,000& (36.8)&1,000 &(45.8) \\
\midrule
\multirow{3}{*}{{EmailSum}} &Train & 1,800 & 6.5 & 231.3 & 26.9& 2,581& (33.2)&730& (33.8)&3,939 &(33.0)&3,203 &(29.7)&7,547& (24.4)&0& (-)\\
&Dev. & 249 & 6.5 & 227.2 & 26.2 & 498 &(37.8)&498& (36.6)&498 &(36.1)&498& (34.0)&498& (24.8)&498 &(35.9) \\
&Test & 500 & 6.5 & 243.0 & 28.2 & 1,000 &(37.0)&1,000 &(36.2)&1,000 &(35.3)&1,000& (32.4)&1,000& (25.7)&1,000& (35.2) \\
\midrule
\multirow{3}{*}{{QMSum}} &Train & 1,095 & 52.6 & 1,137.4 & 71.2& 2,973 &(38.3)&624 &(37.2)&2,197 &(31.7)&2,617 &(29.9)&2,539 &(29.5)&0 &(-)\\
&Dev. & 237 & 57.7 & 1,145.4 & 71.4 & 474 &(36.0)&474 &(33.9)&474 &(33.0)&474 &(28.1)&474 &(29.5)&474 &(24.9) \\
&Test & 244 & 55.6 & 1,152.2 & 63.9 & 488 &(37.4)&488 &(35.0)&488 &(33.8)&488 &(29.4)&488 &(29.1)&488 &(24.6) \\
\midrule
\multirow{3}{*}{{TweetSumm}} &Train & 879 & 10.5 & 244.0 & 48.2& 678 &(47.3)&649 &(47.2)&919 &(43.2)&3,901 &(30.7)&2,643 &(34.9)&0 &(-)\\
&Dev. & 110 & 10.2 & 226.1 & 48.4 & 220 &(52.6)&220 &(50.0)&220 &(48.7)&220 &(34.8)&220 &(35.4)&220 &(49.0) \\
&Test & 110 & 10.6 & 258.2 & 47.8 & 220 &(48.4)&220 &(46.9)&220 &(44.4)&220 &(32.6)&220 &(36.1)&220 &(45.4) \\
\bottomrule[1pt]
\end{tabular}}
\end{center}
\vspace{-7pt}
\caption{\label{tb:dataset}{Statistics of the \textsc{Olds} dataset. \textsc{Olds} is built upon five dialogue summarization benchmarks that cover different domains. \textbf{Len.} stands for the average length (number of words). $\mathrm{L,B,S}$ in the subscript of model names stand for {\em large}, {\em base}, and {\em small} model sizes.}}
\end{table*}

\subsection{The Definition of Omission}


In summarization tasks, omission\footnote{In some previous works, omission is also called missing information \cite{chen2020multi,tang-etal-2022-confit}.} is one of the most common factual errors in abstractive summaries. It usually refers to the missing content in the candidates, which is presented in the gold reference. The definition of omission content is flexible, which could refer to either the omitted keywords, text spans, or utterances. In dialogues, an utterance has a clearer boundary compared to text spans and can be viewed as a basic unit for identification and evaluation. Therefore, in this paper, we mainly focus on {\bf utterance-level omission} and provide utterance-level labels. Table~\ref{tb:example} shows an example of our \textsc{Olds} dataset, which contains the original dialogue, reference summary, candidate summary, and omission labels. In this example, the candidate summary omits three key messages: the person ``\ul{Adam}'', the attitude ``\ul{worried}'' and the persona ``\ul{psychologist}'', and thus the corresponding utterance-level omission labels are the 3rd and 15th utterances in the original dialogue. 

\subsection{Dataset Creation}
\textsc{Olds} is a dataset that collects multiple candidates for dialogue summarization and provides their corresponding omission labels at the utterance level. This dataset first collects multiple public benchmarks, including SAMSum~\cite{gliwa2019samsum}, DialogSum~\cite{chen2021dialogsum}, EmailSum~\cite{zhang2021emailsum}, QMSum~\cite{zhong2021qmsum} and TweetSumm~\cite{feigenblat2021tweetsumm}, to cover different dialogue domains. 

Then, in order to collect samples with omission contents, we still need to generate candidate summaries for each dialogue. To gain deeper insights into the omission problem induced by models with different capacities, we select 6 different model settings~\footnote{We do not use extractive models because dialogue summaries are very abstractive. There is a huge gap in the format and style between summary sentences and dialogue utterances.}, including BART$_{\mathrm{large/base}}$~\cite{lewis2020bart}, T5$_{\mathrm{base/small}}$~\cite{raffel2020exploring}, Transformers~\cite{vaswani2017attention}, and Pegasus$_\mathrm{large}$~\cite{zhang2020pegasus}, to generate candidate summaries~\footnote{Pegasus$_\mathrm{large}$ is only used to generate summaries for dialogues in the validation/test sets. The purpose is to conduct the robustness evaluation on candidates from unseen sources. }.

Finally, based on the collected candidate summaries, we need to identify which salient information is omitted in these candidates. Therefore, we elaborately design a strategy to label omission automatically and the details are described in the next subsection. As a result, our \textsc{Olds} is able to obtain multiple candidates and their corresponding omission label for each dialogue. More details about the dataset creation can refer to Appendix~\ref{sec:appendix}.

\subsection{The Automatic Labeling Strategy}
\label{sec:labeling}
It is generally a non-trivial task to identify the missing critical content in candidate summary. Fortunately, the existing datasets provide reference summaries as ground truths. We could locate the omitted information in dialogue by directly comparing candidates with references. Thus, we design a pipeline strategy for automatic omission labeling, which is composed of three steps: oracle extraction, omission identification, and redundancy removal. Appendix~\ref{appendix:labeling} shows an example of the complete process of automatic omission labeling.

\paragraph{Oracle Extraction}  The first step is to match summaries to the corresponding utterances in the dialogue. Following Nallapati et al. \shortcite{nallapati2017summarunner}, we use a greedy algorithm to select utterances from the dialogue that maximizes the Rouge score~\cite{lin2004rouge} with respect to the summary. We return this subset of utterances as oracle labels, representing their membership in the summary. We define the extracted oracle labels for reference summaries and candidate summaries as {\em Gold Oracle} and {\em Candidate Oracle}, denoted as $G$ and $C$ respectively.

\paragraph{Omission Identification} The goal of this step is to find out the omission set $O$. An intuitive solution is to calculate the complement of candidate oracle in gold oracle as $G-C= \{u|u\in G,u\notin C\}$. Nevertheless, it is an imperfect solution because the utterances in $C$ might still contain omitted words or phrases. For instance, in Table~\ref{tb:example}, the 15th utterance with a phrase {\em ``I have a friend who's a psychologist''} matches the key information ``{\em friend}'' in both reference and candidate, and this utterance would be included in both $G$ and $C$. However, the keyword ``{\em psychologist}'' is actually omitted in the candidate, so the 15th utterance should be labeled as an omission. In other words, some utterances in the intersection of $G$ and $C$ may also be omissions. To further discover the potential omission utterances from $G\cap C=\{u|u\in G,u\in C\}$, we empirically adopt a word-level comparison approach. Specifically, for each utterance $u$ in $G\cap C$, we further extract the overlapping words $W_G^u$ / $W_C^u$~\footnote{We process words in a case-insensitive setting. We keep the original form of words but perform word stemming for comparison. Besides, stop words are removed.} between $u$ and reference/candidate summary. If $W_G^u\not\subseteq W_C^u$, we deem this corresponding utterance includes some key messages that are omitted in the candidate, and thus it should be labeled as an omission. During this process, we could obtain the omission words of utterance $u$, which is denoted as $W^u=\{w|w\in W_G^u, w\notin W_C^u\}$. 

\paragraph{Redundancy Removal} After the omission identification, we can obtain the omission set $O$. However, some utterances in $O$ can be redundant since they could share the identical missing content. For example, for utterance $u_1$ and $u_2$, their omission words $W^{u_1}$ and $W^{u_2}$ can be equal so that we can argue these two utterances share similar omission information. To reduce this redundancy, we only keep the utterance with the front position if multiple utterances have the same omission words.

\begin{table}[t!]
\fontsize{8.5pt}{9.5pt}\selectfont
\begin{center}
\begin{tabular}{lcc}
\toprule[1pt]
\bf Domain &\bf Avg. Accept Num. (Rate)& \em \textbf{kappa} \\
\midrule
SAMSum & 182.3$_{\pm 3.6}$ (91.2\%) & 0.689\\
DialogSum & 188.0$_{\pm 4.4}$ (94.0\%) & 0.616 \\
EmailSum & 192.3$_{\pm 2.5}$ (96.2\%) & 0.633\\
QMSum& 197.0$_{\pm 1.0}$ (98.5\%)& 0.549 \\
TweetSumm& 194.0$_{\pm 1.7}$ (97.0\%)& 0.656\\
\midrule
Overall& 953.7$_{\pm 4.2}$ (95.4\%)& 0.653 \\
\bottomrule[1pt]
\end{tabular}
\end{center}
\vspace{-6pt}
\caption{\label{tb:assess} Quality assessment based on human evaluation. We randomly sampled 200 examples for each domain and asked 3 annotators to rate {\em Accept} or {\em Reject}.}
\end{table}

\subsection{Quality Assessment}
To assess the quality of the extracted omission labels for the \textsc{Olds} dataset, we also conducted human evaluation to validate the correctness of the labeled utterances. We recruited three annotators with NLP backgrounds and each annotator is required to answer the question whether the set of labeled omission utterances is {\em Accept} or {\em Reject}. The set should be marked as {\em Reject} as long as it misses any critical utterance (recall of labeled omissions), or includes any redundant or uninformative utterance (precision of labeled omissions). Otherwise, it should be marked as {\em Accept}. To this end, we randomly sampled 200 dialogue-candidate pairs from each domain for assessment. Table~\ref{tb:assess} reports the results of the human evaluation for quality assessment. The acceptance rate of human evaluation ranges between 91.2\%-98.5\%, which validates the effectiveness of our omission extraction strategy. Furthermore, in order to evaluate the reliability of this assessment, we measure the agreement between different annotators by reporting Fleiss' Kappa values~\cite{fleiss1971mns} among the possible combinations of two annotators, as reported in Table~\ref{tb:assess}. We find that the overall Kappa score is 0.653, which shows the substantial agreement between annotators. Overall, the results of human evaluation demonstrate that our omission extraction strategy is able to produce high-quality omission labels automatically. More details about human evaluation can refer to Appendix~\ref{appendix:human_evaluation}.

\subsection{Dataset Format and Statistics}
An example of our \textsc{Olds} dataset is shown in Table~\ref{tb:example}, which contains the basic information, such as dialogue, reference, candidate, and omission labels. In the released version of \textsc{Olds}, we further provide some auxiliary information. The detailed dataset format and a complete example can be seen in Appendix \ref{appendix:dataset_format}. Table~\ref{tb:dataset} shows the statistics of the \textsc{Olds} dataset.
We can see that the dialogues are from different domains, with different lengths and turns. Besides, the lengths of summaries also differ from each other, and the employed abstractive models are able to produce candidates with different qualities. We expect that our dataset could pave the way for analyzing the omission problem across different domains and diverse candidate summaries.



\section{Understanding the Omission Problem}
In this section, we explore the omission problem in different aspects and analyze why we should pay attention to omission in dialogue summarization.

\subsection{Distribution of Omission Information}
To explain the importance of the omission problem, we answer the following two questions.

\begin{figure}
\centering
  \includegraphics[width=2.6in]{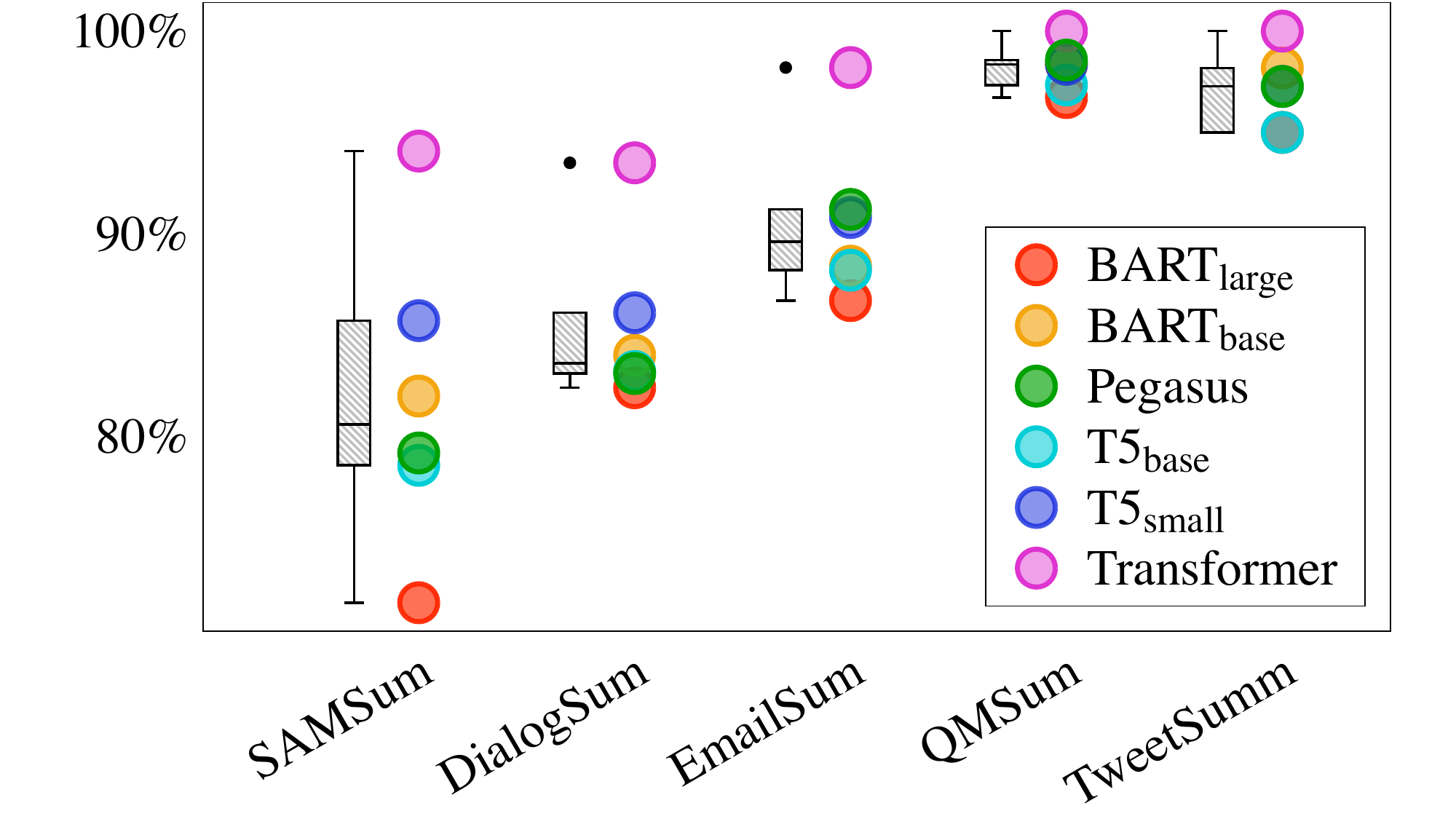}
  \vspace{-7pt}
  \caption{The percentage of candidate summaries with omission errors. We report the results of six adopted models on the test set of each dialogue domain. }  
  \label{fig:model-dist}
\end{figure}

\paragraph{Q1: How serious is the omission problem?} For each abstractive model used in \textsc{Olds}, we calculate the percentage of candidates which include omission information (i.e., the omission set $O \neq \emptyset$). 
Generally, a lower percentage means the model's ability to identify the salient information in dialogue is more powerful. Figure~\ref{fig:model-dist} shows the statistical results of each model on different dialogue domains. We find that using pre-trained models always produces a lower ratio than the vanilla Transformer. Nevertheless, even using pre-trained models, we find it still reaches a high omission ratio of at least 70\%. The omission phenomenon is worse in QMSum and TweetSumm, that almost 90\% of their candidates have omission errors. From this perspective, we can conclude that omission is a general and grievous problem in dialogue summarization, and how to alleviate the omission problem is still intractable.
\begin{table}[t!]
\fontsize{8.0pt}{9.5pt}\selectfont
\vspace{-2pt}
\begin{center}
\begin{tabular}{lccccc}
\toprule[1pt]
\bf Domains&\bf SAM.  & \bf Dial.&\bf Email. &\bf QM. &\bf Tweet. \\
\midrule
\bf RG-1 & -0.563& -0.409&-0.445 &-0.470 &\bf -0.574\\
\bf RG-2 & -0.448& -0.342& -0.394& -0.480& -0.524\\
\bf RG-L & -0.510& -0.398&\bf -0.457& -0.494& -0.547\\
\bf BLEU &-0.332 & -0.289& -0.231& -0.397& -0.467\\
\bf BS$_\mathrm{B}$ &-0.549 & -0.502& -0.418& -0.463& -0.485\\
\bf BS$_\mathrm{L}$ &-0.562 &\bf -0.504& -0.445&\bf -0.521& -0.546\\
\bf BLEURT &\bf -0.567& -0.461& -0.292& -0.410& -0.525\\
\bottomrule[1pt]
\end{tabular}
\end{center}
\vspace{-5pt}
\caption{\label{tb:correlation} Pearson correlations between {\em Omission Rate} and other reference-based metrics on the test set of five domains. {\bf RG} denotes ROUGE. {\bf BS$_\mathrm{B}$}, {\bf BS$_\mathrm{L}$} stand for BERTScore using {\em Roberta-base} and {\em Roberta-large} as backbone models. For BLEURT, we use {\em BLEURT-20}.}
\end{table}

\begin{figure}
\centering
  \includegraphics[width=3.0in]{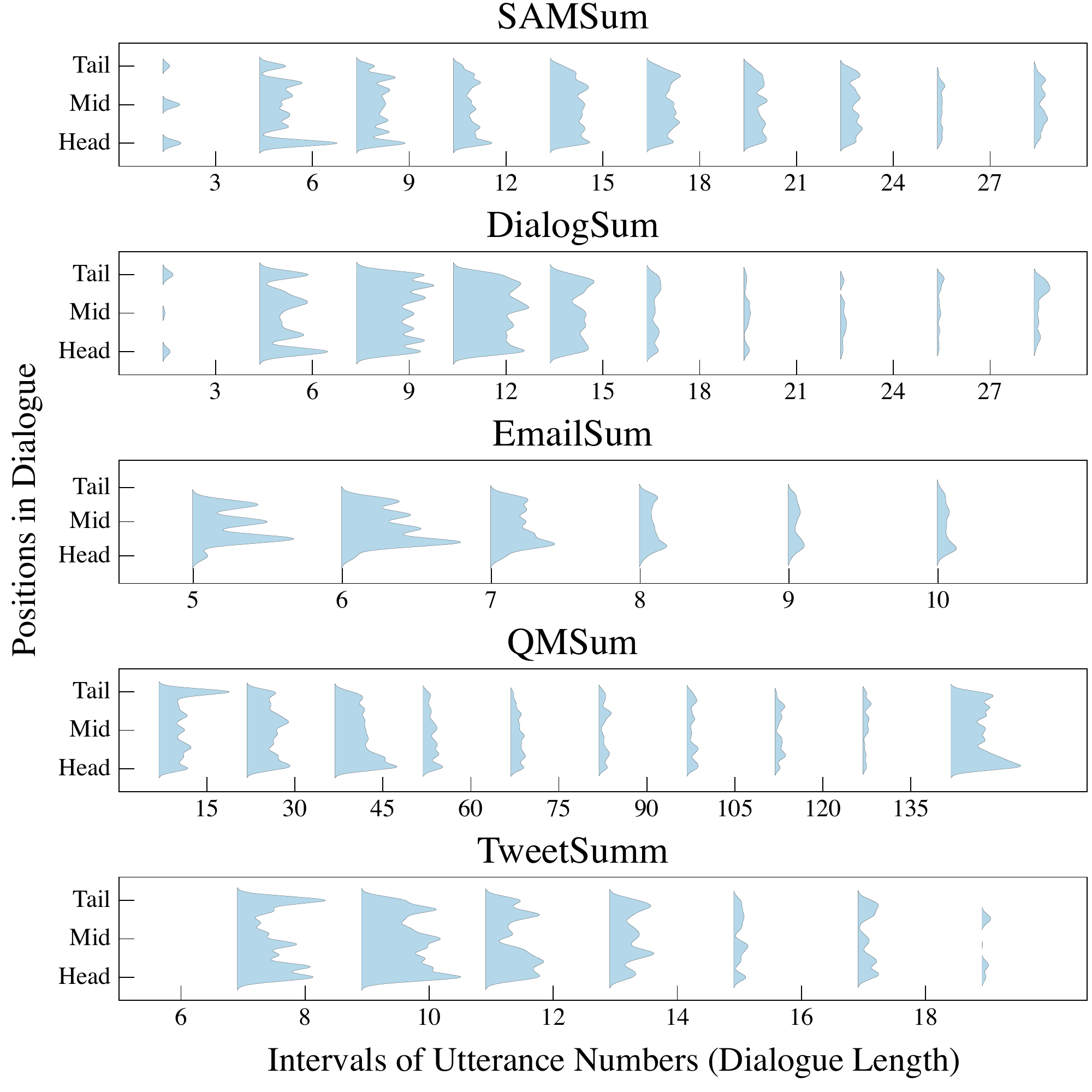}
  \vspace{-10pt}
  \caption{Position distribution of omissions in dialogues across different domains. The X-axis represents the intervals of untterance numbers. }  
  \label{fig:dialog-dist}
\end{figure}

\paragraph{Q2: How is the omission information distributed in the dialogue? } 
To answer this question, we investigate the position distribution of omissions in dialogues. Just as shown in Figure~\ref{fig:dialog-dist}, we observe that the omitted utterances are randomly distributed in each position of the dialogue, regardless of its length and domain. This position distribution also indicates that dialogues are unstructured, and how to identify the dispersed key information precisely is still difficult for current models.

\subsection{Correlation with Reference-Based Metrics}
Since omission is defined by the difference between references and candidates, we thus investigate the correlation between the amount of omission content and a variety of reference-based metrics, to verify whether the omission rate of a candidate summary could affect these metrics. Here, we calculate the {\em omission rate} as follows:
\begin{equation}
    \label{eq:omission_rate}
    {\rm OmissionRate} = \frac{\sum_{u\in O}|W^u|}{\sum_{u\in G}|W^u_G|},
\end{equation}
where $W^u$ and $W^u_G$ denote the set of omitted words and the set of gold oracle words shared across $u$ and the reference, respectively. It directly measures the amount of key information omitted by a summary, and a lower rate indicates the candidate is of higher quality. Table \ref{tb:correlation} demonstrates the Pearson correlations between omission rate and other reference-based metrics, including $n$-gram based metrics ROUGE~\cite{lin2004rouge} and BLEU~\cite{papineni2002bleu}, embedding-based metric BERTScore \cite{zhang2019bertscore}, and learning-based metric BLEURT \cite{sellam2020bleurt}. The results indicate that most of the reference-based metrics moderately correlate with the omission rate, among which BERTScore$_\mathrm{Large}$ is the most stable metric that has a better correlation with the amount of omission content. By contrast, BLEU shows the least correlation because it is a precision-oriented metric. Empirical analyses indicate that the omission rate is strongly correlated with a wide range of evaluation metrics, and so how to mitigate the omission problem is one of the most important priorities to improve the quality of dialogue summaries.

\subsection{Omission-based Summary Refinement}
\label{sec:refinement}
\begin{figure}[!t]
  \centering
  \includegraphics[width=2.9in]{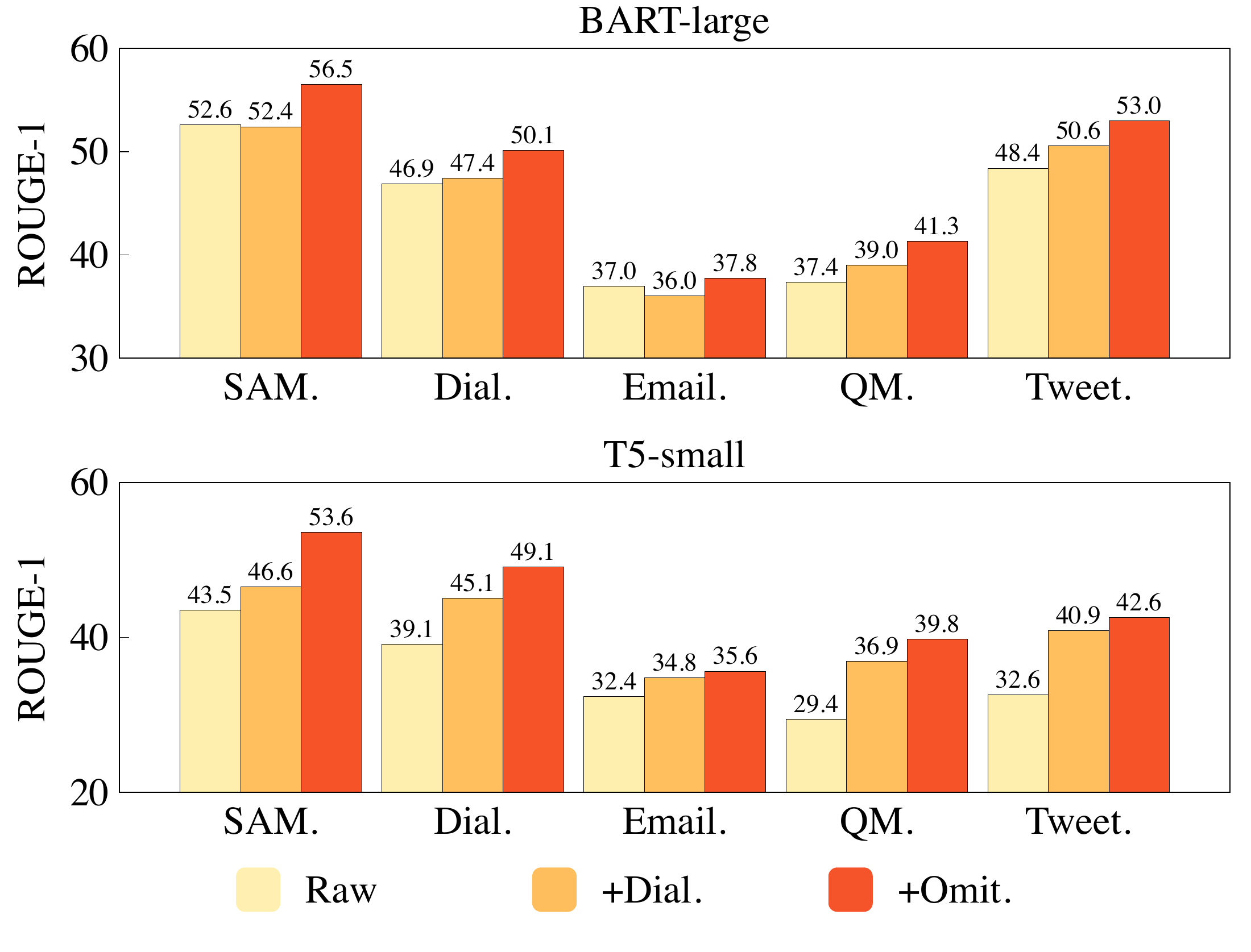}
  \vspace{-5pt}
  \caption{Post-editing results in different domains. {\em Raw} means the results of raw candidates. {\em +Dial.} and {\em +Omit.} mean using raw dialogue or omissions as the supplement information for refinement.}  
  \label{fig:post-edit}
\end{figure}
The above analyses demonstrate the importance of omission information. So we raise another question: what happens if we utilize the omissions to refine the summary quality? Hence, we adopt a post-editing method to investigate the potential of using omissions.
Specifically, we formulate summary refinement as a seq2seq task to predict the gold summary. Instead of inputting raw dialogue, we use the concatenation of candidate summary, omission utterances, and non-omission utterances as the input: ``{\em Candidate} <sep> {\em Omission} <sep> {\em Non-Omission}''. By dividing dialogue utterances into the omission and non-omission groups, the model is able to distinguish omission information while perceiving the whole dialogue simultaneously. If the omission group is empty, it is identical to using candidate and raw dialogue for refinement, and we consider it as the baseline for comparison.  We use BART$_\mathrm{large}$ and T5$_\mathrm{small}$ as the backbone model, and the results are shown in Figure~\ref{fig:post-edit}. The results show that performances are significantly enhanced by the refinement using omissions compared to that using raw dialogues. Notably, on some datasets like SAMSum and DialogSum, T5$_\mathrm{small}$ with supplementary omission information even outperforms the raw BART$_\mathrm{large}$, which indicates that omission-based refinement is a promising direction for quality improvement in dialogue summarization.

\begin{figure}[!t]
  \centering
  \includegraphics[width=2.3in]{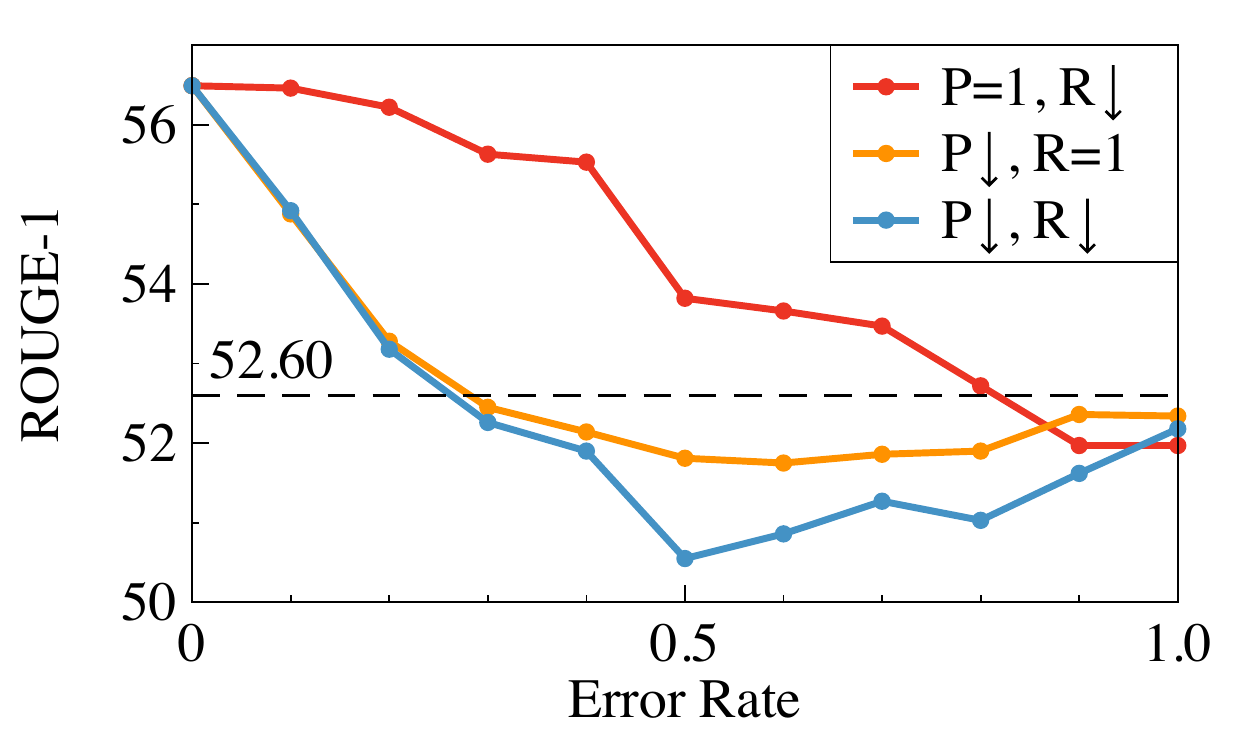}
  \vspace{-5pt}
  \caption{Change of the post-editing results by perturbing input omissions. The results are from the SAMSum test set using BERT$_\mathrm{large}$. The dotted line shows the raw results before post-editing. P and R denote the precision and recall of omissions. $\downarrow$ stands for a decreasing trend.}  
  \label{fig:post-edit-trend}
\end{figure}

In addition, Figure~\ref{fig:post-edit} also shows an upper bound of performance boost by post-editing because we directly employ the gold omission utterances. However, in real situations, we may identify some incorrect omissions. To further explore the impact of wrong omissions on the post-editing results, we investigate three different perturbations by gradually injecting errors into the omission group: 1) we keep the precision as 1 and decrease the recall by moving utterances from the omission group to the non-omission group; 2) we keep the recall as 1 and decrease the precision by moving utterances from the non-omission group to the omission group; 3) we gradually exchange utterances in the two groups until they are swapped, and both the precision and recall decrease from 1 to 0. Figure~\ref{fig:post-edit-trend} depicts the trend of performance degradation as the error rate increases. From the curves, we can find that the precision is relatively more important because the refinement model performs more robustly in the first type of perturbation and is sensitive to the addition of wrong omissions.

\begin{table*}[t]
\fontsize{7.1pt}{9pt}\selectfont
\begin{center}
\setlength{\tabcolsep}{0.7mm}{
\begin{tabular}{lccccccccccccccccccccccccc}
\toprule[1pt]
 \multirow{2}{*}{{\bf Model}}&& \multicolumn{4}{c}{\bf SAMSum} && \multicolumn{4}{c}{\bf DialogSum} && \multicolumn{4}{c}{\bf EmailSum} && \multicolumn{4}{c}{\bf QMSum} && \multicolumn{4}{c}{\bf TweetSumm} \\
  \cline{3-6}
 \cline{8-11}
\cline{13-16}
 \cline{18-21}
\cline{23-26}
 && P & R & F1 & WR && P & R & F1 & WR && P & R & F1 & WR && P & R & F1 & WR && P & R & F1 & WR \\
\hline
\multicolumn{26}{c}{\em Pair-wise Classification} \\
BERT && 41.66 & 38.60 & 40.07 & 54.83 && 38.01 &45.56 &41.44 & 57.23 && 47.94 &40.81 &44.09 &50.93 &&35.97 &42.86&39.12 & 60.73&& 41.84 &47.17& 44.35 & 53.86 \\
RoBERTa && 42.45 & 43.27 & 42.85 & 58.94 && 38.42 & 44.93 & 41.43 & 57.56 && 48.13 & 50.02 &49.05 & 59.04 && 32.92 &\bf 43.65	&37.53 &\bf 60.99 && 41.37 &49.66 &45.14 & 57.08 \\
\hline
\multicolumn{26}{c}{\em Sequence Labeling} \\
BERT &&45.18 &43.57 &44.37 & 61.35 && 40.71 & 46.23 & 43.30 & 57.51 && 50.58 &\bf 50.41 & 50.49 &\bf 61.11 &&47.11&31.22 &37.56 & 47.29&&40.70 &49.72 &44.76 & 58.48 \\
RoBERTa && 47.34 &\bf 47.65 &\bf 47.49 &\bf 63.97 && 42.63 &\bf 46.54 &44.50 &58.51 &&\bf 53.62 &48.65 &\bf 51.01 & 59.04 &&\bf 48.09 & 36.27 &\bf 41.35 & 52.82 && 48.26 &48.85 &48.55 & 59.27 \\
\hline
\multicolumn{26}{c}{\em Pointer Network} \\
BERT && 47.20 &	39.13 &	42.79 & 58.52 && 41.23& 42.79 &42.00 & 56.02 && 52.57 & 48.47 & 50.44 & 60.66 &&45.31&31.73&37.32&48.46&&\bf 48.69&42.35 &45.30 & 52.10\\
RoBERTa &&\bf 50.64 &41.90 &45.86 & 60.25 &&\bf 43.68 &45.90 &\bf 44.76 & \bf 60.03 && 53.61 & 46.04 & 49.54 & 56.92 && 44.80 & 35.16 & 39.40 & 53.21 && 47.23 &\bf 52.18 &\bf 49.58 &\bf 63.61 \\
\bottomrule[1pt]
\end{tabular}}
\end{center}
\vspace{-5pt}
\caption{\label{tb:main_result}{Experimental results of omission detection on \textsc{Olds} dataset. WR means the word-level omission recall.}}
\end{table*}
\section{The Omission Detection Task}
Since candidate summaries could be effectively improved given the gold omission information, how to accurately detect omission utterances in dialogue naturally becomes a critical question. In this section, we formulate the omission detection task in a reference-agnostic setting. Formally, given a dialogue $D=\{u_1,u_2,..,u_N\}$ along with a candidate summary $c$, a detection model is required to extract a set of omission utterances $O$ from $D$ without knowing the reference summary. In this section, we introduce three typical frameworks as baselines and conduct evaluations to see how this task could benefit from them.

\subsection{Model Settings}
To build a foundation for the omission detection task and explore what model architecture the task could benefit from, we investigate three frameworks as baselines, which have different input formats and structures. Their implementation and training details can be found in Appendix~\ref{appendix:detection-setting}.

\paragraph{Pair-wise Classification} A straightforward way is to model this task as an utterance-level classification problem. The input pattern for this paradigm is: {\em <s>} $c$ {\em </s>} $u_i$ {\em </s>}, where {\em <s>} and {\em </s>} denote the classification token and separation token, respectively. $c$ is the candidate summary and $u_i$ is the $i$-th utterance in the dialogue. The model would perform binary classification for the candidate-utterance pair as $y \in \{0, 1\}$, where $y=1$ represents that the utterance is identified as an omission.

\paragraph{Sequence Labeling} Inspired by BERTSum~\cite{liu2019text} that formulates extractive summarization as a sequence labeling problem at the sentence level, we employ a similar strategy which assigns each utterance a label $y_i \in \{0, 1\}$ indicating whether the utterance is an omission. We append the candidate summary in front of the dialogue, as {\em <s>} $c$ {\em </s>} {\em <s>} $u_1$ {\em </s>} {\em <s>} $u_2$ {\em </s>} ... {\em <s>} $u_N$ {\em </s>}. The last hidden layer of each {\em <s>} token will be used as utterance representations for classification.

\paragraph{Pointer Network} Pointer network is to select the omission utterance recurrently using glimpse operation~\cite{vinyals2015order} based on previous predictions. It is a widely-used strategy for sentence extraction in summarization~\cite{chen2018fast,zhong2019searching,zou2021topic}. Here, we use the same input format as in sequence labeling, and the pointer network outputs an extraction distribution based on the {\em <s>} representations.

\subsection{Evaluation Metrics}
We use the standard Precision (P), Recall (R), and F1-score (F1) metrics on the utterance level to evaluate omission detection models. Furthermore, we calculate the percentage of gold omission words that are hit in the detected utterances to measure the word-level omission recall:
\begin{equation}
    WR = \frac{\# \mathrm{hit\ omission\ words}}{\# \mathrm{gold\ omission\ words}}.
\end{equation}
$\#$ means the counted number. The closer the word-level omission recall is to 1, the more the omission information is collected by the detection model.

\begin{table*}[t]
\fontsize{7.8pt}{9pt}\selectfont
\begin{center}
\setlength{\tabcolsep}{0.7mm}{
\begin{tabular}{clccccccccccccccccccccc}
\toprule[1pt]
 \multirow{2}{*}{{\bf \small Framework}}&\multirow{2}{*}{{\bf \small Model}}&& \multicolumn{2}{c}{\bf Overall} && \multicolumn{2}{c}{\bf BART$_\mathrm{Large}$} && \multicolumn{2}{c}{\bf BART$_\mathrm{Base}$} && \multicolumn{2}{c}{\bf T5$_\mathrm{Base}$} && \multicolumn{2}{c}{\bf T5$_\mathrm{Small}$} && \multicolumn{2}{c}{\bf Transformer} && \multicolumn{2}{c}{\bf Pegasus$_\mathrm{Large}$}\\
 \cmidrule{4-5}
 \cmidrule{7-8}
\cmidrule{10-11}
 \cmidrule{13-14}
\cmidrule{16-17}
 \cmidrule{19-20}
 \cmidrule{22-23}
& && F1 & WR & & F1 & WR && F1 & WR && F1 & WR && F1 & WR && F1 & WR && F1 & WR\\
\midrule
\multirow{2}{*}{{\em Pair-wise classification}} & BERT && 40.07 & 54.83 & &31.98 & 54.77 && 38.40 & 55.14 && 33.80 & 51.75 && 42.83 & 55.51 && 48.60 & 56.69 && 38.25 & 55.11 \\
& RoBERTa && 42.85 & 58.94 && 35.52 & 58.32 && 41.28 & 57.86 && 38.75 & 58.03 && 44.68 & 59.80 && 51.43 & 61.29 && 39.39 & 58.36 \\
\multirow{2}{*}{{\em Sequence Labeling}}& BERT && 44.37 & 61.35 && 35.58 & 59.94 && 42.23 & 60.23 && 40.27 & 60.63 && 46.70 & 62.77 && 53.82 & 64.85 && 41.07 & 59.60 \\
& RoBERTa &&\bf 47.49 &\bf 63.97 &&\bf 38.37 &\bf 61.23 &&\bf 45.66 &\bf 62.51 &&\bf 43.50 &\bf 62.42 &&\bf 50.41 &\bf 66.81 &&\bf 55.94 &\bf 67.43 &&\bf 44.67 &\bf 63.32 \\
\multirow{2}{*}{{\em Pointer Network}}& BERT && 42.79 & 58.52 && 35.75 & 58.53 && 39.82 & 57.65 && 38.86 & 57.83 && 44.91 & 59.32 && 52.04 & 60.55 && 39.57 & 57.22 \\
& RoBERTa && 45.86 & 60.25 && 37.12 & 58.27 && 43.54 & 58.81 && 40.88 & 58.01 && 48.06 & 62.50 && 54.78 & 63.90 && 44.03 & 60.03 \\
\bottomrule[1pt]
\end{tabular}}
\end{center}
\vspace{-6pt}
\caption{\label{tb:source}{Omission detection results on the candidate summaries from SAMSum test set, which are categorized into multiple groups according to their source summarizer.}}
\end{table*}

\subsection{Main Results}
Table~\ref{tb:main_result} presents the experimental results on \textsc{Olds}. All detection models are separately trained on the five domains. For each omission detection framework, we employ BERT$_\mathrm{base}$ and RoBERTa$_\mathrm{base}$ as the backbone model to extract text features. Among these three frameworks, pair-wise classification performs the worst in most cases since it does not consider contextual information of dialogue. Meanwhile, sequence labeling is on par with the pointer network, which indicates that dialogue context is a crucial factor for models to detect the omitted content. However, although omission detection models only need to make a choice of whether the given utterance is an omission, the task is still very challenging. In Table~\ref{tb:main_result}, the best F1 score is around 50\% in all five domains, while the recalled omission words in extracted utterances (WR) are around 60\%. Besides, models in QMSum only achieve at most a F1-score of 41.35 and we guess it is due to the effect of longer dialogue in QMSum (over 1K tokens in Table~\ref{tb:dataset}). Intuitively, summarizers produce the candidates that have picked the low-hanging fruit, and the remaining omission information is a tough nut to crack. In other words, there exists some salient information omitted by the summarizer that is still difficult for detection models to capture.

\begin{figure}
\centering
  \includegraphics[width=3.0in]{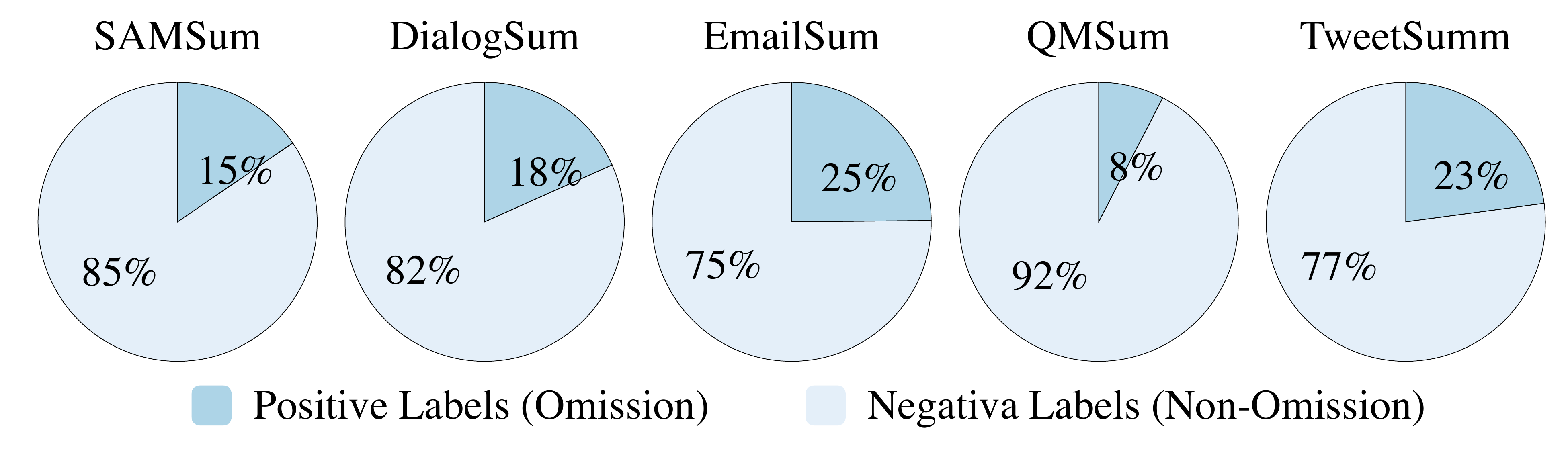}
  \caption{The proportion of positive labels (omission utterances) against negative ones (non-omission utterances) in five domains.}  
  \label{fig:imbalance}
\end{figure}



\subsection{Analysis and Discussion}
To understand what factors may affect the performance of the detection model, we conduct the following explanatory experiments.

\paragraph{Label Imbalance} We first calculate the percentage of omission utterances against non-omission utterances in five domains to investigate whether the label imbalance problem exists in the datasets. Figure~\ref{fig:imbalance} shows that the proportion of positive labels is always smaller than 25\%, which indicates that label imbalance is a common problem in omission datasets. Besides, we observe that the degree of label imbalance is consistent with the performance of detection models, according to the results in Table~\ref{tb:main_result}. For example, the models achieve nearly 50\% F1-score in EmailSum and TweetSumm, which have a ratio of 25\% and 23\% omission utterances. However, in QMSum, the detection models only achieve a 40\% F1-score as the omission proportion of this dataset is only 8\%. Hence, how to alleviate label imbalance is critical for omission detection and we leave it as future work.

\paragraph{Candidate Quality} Furthermore, we evaluate the performance of detection models on the candidates produced by different abstractive summarizers to investigate whether the candidate quality may influence detection models. The results are shown in Table~\ref{tb:source}, and we find the result of omission detection is negatively correlated with the performance of summarizers. For instance, BART$_\mathrm{L}$ and Pegasus$_\mathrm{L}$ produce candidates with higher quality, yet the detection model has difficulty obtaining their omissions. On the contrary, Transformer produces relatively low-quality candidates, while the detection model could produce better results (i.e., 55.94\% F1-score). It indicates that capturing the remaining omissions for high-quality candidates is difficult, and how to address this issue is also valuable.

\paragraph{Cross-Domain Results} In addition, we conduct the cross-domain evaluation to investigate domain gaps and the generalizability of detection models. From Table~\ref{tb:cross_domain}, we can conclude that there are obvious differences between these five domains. For example, the models trained on the other domains perform poorly when tested directly on QMSum. Among these five domains, the difference between SAMSum and DialogSum is relatively small due to their similar performances across domains. We also find that the model trained on the large dataset SAMSum has a better capability of generalizing to other domains, even achieving the best result on the small datasets DialogSum and EmailSum. 

\begin{table}[t!]
\fontsize{8pt}{9.5pt}\selectfont
\begin{center}
\begin{tabular}{lccccc}
\toprule[1pt]
\bf Domains&\bf SAM.  & \bf Dial.&\bf Email. &\bf QM. &\bf Tweet. \\
\midrule
\bf SAM. &\bf 63.97&\bf 59.39 &\bf 66.80 & 36.68 & 53.92\\
\bf Dial. & 49.65 & 58.51& 66.58& 43.50& 53.77\\
\bf Email. & 37.78 & 30.69 &59.04& 20.98 & 24.13\\
\bf QM. & 41.39 & 47.00& 61.15 &\bf 52.82& 28.20\\
\bf Tweet. & 44.92& 48.46& 57.07& 14.74 &\bf 59.27\\
\bottomrule[1pt]
\end{tabular}
\end{center}
\caption{\label{tb:cross_domain} Cross-domain evaluation results. Each row represents the training set, and each column represents the test set. We use the sequence labeling framework equipped with RoBERTa$_\mathrm{base}$ for these experiments and use the word-level omission recall (WR) for evaluation. }
\end{table}
\subsection{Future Research Opportunities}

From the results in Table~\ref{tb:main_result}, we could observe that omission detection is a challenging task. Hence, we summarize some research directions as follows:
\vspace{-18pt}
\begin{itemize}[leftmargin=*]
    
    \setlength{\itemsep}{2pt}
    \setlength{\parsep}{2pt}
    \setlength{\parskip}{2pt}
    \item One direction is to develop a more advanced model for omission detection. Based on the analysis of Section~\ref{sec:refinement}, we could focus on improving the precision of omission detection results because a high precision of detected omissions would bring benefit to the refinement model. An ideal detection model could serve as a model-based metric for reference-free summary evaluation. Besides, we could use the detected omission to improve the results of summarization. 
    \item Another research direction is to develop a refinement model for summary improvement using the detected omissions. In this paper, we briefly touch on this by introducing a post-editing approach in Section~\ref{sec:refinement}. The approach is straightforward, and the whole summarization procedure becomes a summarize-then-refine pipeline. However, the results show that the model is sensitive to wrong omissions. Hence, how to design a robust refinement model is also noteworthy.
\end{itemize}

\section{Related Work}
\subsection{Dialogue Summarization}
Dialogue summarization is a challenging and valuable task that has recently received much attention, where a variety of dialogue domains are investigated, such as mail threads \cite{rambow2004summarizing,zhang2021emailsum}, meetings \cite{chen2020multi,zhong2021qmsum}, customer service \cite{,zou2021unsupervised,zou2021topic,feigenblat2021tweetsumm}, medical conversations \cite{joshi2020dr,song2020summarizing}, and daily chats \cite{gliwa2019samsum,chen2021dialogsum}. Different from conventional documents, dialogues have several inherent characteristics that make the summarization task more challenging \cite{zou2021low,feng2022survey}, e.g., multi-party information, coreferences, topic drifting, etc. Recent works have explored the types of errors in generated dialogue summaries to develop robust models \cite{tang-etal-2022-confit}, and omission is assessed as the most dominant error type in candidate summaries, which is also supported by human evaluations in previous works \cite{chen2020multi,liu2021controllable,liu2021coreference}. In this work, we comprehensively analyze the omission problem in dialogue summarization based on the curated benchmark and investigate the feasibility of omission detection for generated candidates. 

\subsection{Omission in Text Generation Tasks}
Omission is a common error in machine translation (MT)~\cite{russell1999errors,Sharma2015Omission,Yang2019ReducingOmission} and automatic speech recognition (ASR) tasks~\cite{Yue2020ASRCorrection}, which usually denotes the missing source information in the generated sequences. Although both summarization and MT/ASR belong to generation tasks, the definitions of omission error are different among these tasks. In MT/ASR tasks, the tokens between source and target sequences are usually well aligned, which means each token in the target sequence can locate its corresponding content in the source sequence. Due to such characteristics, previous works~\cite{Zhaopeng2016Coverage} in MT/ASR tasks usually adopted coverage mechanisms to eliminate the influence of omission error. Nevertheless, the source sequences in summarization tasks usually include abundant redundant and useless information, especially in dialogue scenarios, which makes omission a more serious problem in summarization-like tasks.

\section{Conclusion}
In this work, we systematically study the omission problem in dialogue summarization based on the curated \textsc{Olds} dataset, which collects candidate summaries from multiple models and domains and provides high-quality omission labels for them. We discover that omission is a significant problem that directly affects the results of dialogue summarization, and the defective candidate summary could be largely improved by leveraging the omission information properly. We further introduce an omission detection task to identify omission content, which is a challenging and valuable task that paves the way to omission mitigation and summary improvement in dialogue summarization. 

\section{Limitations}
The omission problem is critical in dialogue summarization, but even if this problem is solved, we still cannot guarantee a candidate is appropriate because it might bring hallucination content that is not presented by the source dialogue. Previous works~\cite{tang-etal-2022-confit,maynez2020faithfulness} also concluded that factual inconsistency is a critical problem in dialogue summarization, and it is not easy to distinguish. How to mitigate the omission problem while avoiding the occurrence of new errors is not discussed in this paper, and we hope to address this issue in future work.

\section*{Acknowledgments}
The authors wish to thank the anonymous reviewers for their helpful comments. This work was partially funded by National Natural Science Foundation of China (No.62206057), Shanghai Rising-Star Program (23QA1400200), and Natural Science Foundation of Shanghai (23ZR1403500).

\bibliography{anthology,custom}
\bibliographystyle{acl_natbib}

\appendix

\section{Details of the \textsc{Olds} dataset}
\label{sec:appendix}

\subsection{Example of Automatic Omission Labeling}
\label{appendix:labeling}
\begin{figure*}
\centering
  \includegraphics[width=5.0in]{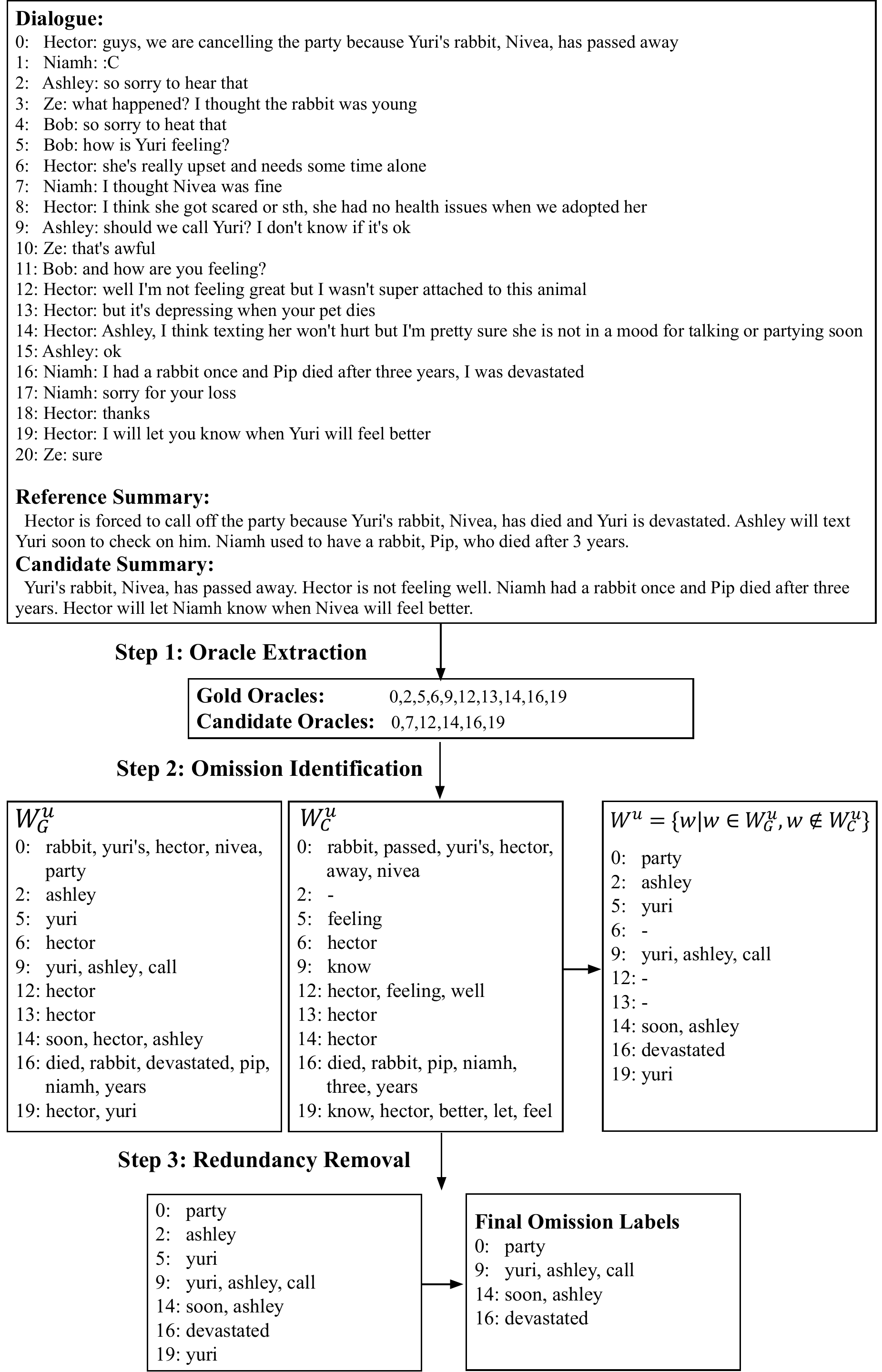}
  \caption{An example of the complete process of automatic omission labeling, which is sampled from the training set of SAMSum. $W_G^u$ is a word set that contains all overlapping words between $u$ and the reference summary. Similarly, $W_C^u$ contains overlapping words between $u$ and the candidate summary. $W^u$ is the set of omission words.}  
  \label{fig:label_process}
\end{figure*}

Figure \ref{fig:label_process} shows an example of the complete process of automatic omission labeling, which consists of three steps: oracle extraction, omission identification, and redundancy removal. 

For oracle extraction, we select utterances greedily from the dialogue to maximize the Rouge score with respect to the summary. We obtain this subset of utterances as oracle labels, representing their membership in the summary. In this example, we generate oracle labels for the reference as {\em Gold Oracles}, i.e., an utterance set of \{0, 2, 5, 6, 9, 12, 13, 14, 16, 19\}, and oracle labels for the candidate as {\em Candidate Oracles}, i.e., \{0, 7, 12, 14, 16, 19\}. 

In the process of omission identification, we traverse the utterances in Gold Oracles and extract $W_G^u$, which is a set of words containing the overlapping words between $u$ and the reference. For instance, in the 14th utterance, "soon, Hector, Ashley" are the keywords appearing in the reference. Similarly, we extract $W_C^u$ that contains the overlapping words between $u$ and the candidate summary, where $u\in$ Gold Oracles. Then, by comparing $W_G^u$ and $W_C^u$, we could obtain the omission words $W^u=\{w|w\in W_G^u, w\notin W_C^u\}$. For any utterance $u$ where $W^u\neq \emptyset$, we label it as an omission utterance. In the example of Figure~\ref{fig:label_process}, the 14th utterance contains the keywords "soon, Ashley" which are omitted by the candidate, and it should be labeled as an omission.

Finally, we conduct redundancy removal to discard redundant omission utterances. In Figure~\ref{fig:label_process}, the 2nd, 5th, and 19th utterances have redundant omission words $W^u$, which are the same as those in other omission utterances. Hence, we remove these utterances and the final omission labels are \{0, 9, 14, 16\}.

\subsection{Dialogue Domains}
\label{appendix:dataset}
We build the \textsc{Olds} dataset upon five existing dialogue summarization datasets that cover different domains, which are described as follows:

\paragraph{SAMSum} It is the first high-quality online chat summarization corpus~\cite{gliwa2019samsum}, which contains about 16k simulated conversations created by linguists with corresponding summaries.

\paragraph{DialogSum} It is a summarization dataset~\cite{chen2021dialogsum} with 13.5k real-life scenario dialogues, which are face-to-face spoken dialogues that cover a wide range of daily-life topics.

\paragraph{EmailSum} It is an email thread summarization dataset~\cite{zhang2021emailsum} that consists of 2,549 email threads along with annotated summaries. The dataset has two types of summaries, short summary ($<$30 words) and long summary ($<$100 words). Here, we use the short version as references because they are more abstractive and challenging.

\paragraph{QMSum} It is a query-based multi-domain meeting summarization benchmark~\cite{zhong2021qmsum} that contains 1,808 query-summary pairs over 232 meetings. We concatenate queries with their corresponding text spans as the input dialogues~\footnote{We removed 232 query-summary pairs which summarize the whole meeting transcripts because their input lengths are significantly different from other pairs. As a result, the final number of pairs used in our dataset is 1,576.}.

\paragraph{TweetSumm} It is a dataset focused on customer service conversations~\cite{feigenblat2021tweetsumm}, which contains 1,100 dialogues, each accompanied by 3 extractive and 3 abstractive summaries. We use the longest abstractive summary as the gold reference.

\subsection{Candidate Generation}
\label{appendix:candidate_gen}
We use 6 abstractive models to generate candidates for the dialogues in \textsc{Olds}, including BART$_{\mathrm{large/base}}$, T5$_{\mathrm{base/small}}$, vanilla Transformer, and Pegasus$_\mathrm{large}$. Pegasus$_\mathrm{large}$ is only used to generate candidates for dialogues in evaluation sets. 

To obtain the candidate summaries in training sets, we train the summarization models by adopting a 10-fold cross-validation approach, and each model generates 10 candidates for each dialogue in the validation fold via different configurations of beam search and sampling. As a result, we can obtain 50 candidates (5 models $\times$ 10 inferences) for each dialogue in the training set. To ensure the diversity of the generated candidates, we further calculate the average Levenshtein distance~\cite{Levenshtein1965BinaryCC} for each candidate and pick out 10 candidates with the largest scores. Specifically, we combine these candidates in pairs (a total of 50 $\times$ 50 = 2,500 pairs) and calculate the Levenshtein distance between them. Then, for each candidate, we average the distance results against the other 49 candidates to obtain the average Levenshtein distance. Finally, we rank these candidates based on the scores in descending order and pick out the top 10 candidates. As a result, we have 10 diverse candidates for each dialogue in the training sets.

For the evaluation set of \textsc{Olds}, we train the aforementioned 6 models on the training set of each domain to produce candidate summaries. Each summarization model produces 2 candidates, which are decoded by beam search (beam size = 5) and sampling, respectively. Hence, we totally have 12 candidates for each dialogue in evaluation sets. 

The training and inference process was conducted based on the official code of pre-trained language models~\footnote{\url{https://huggingface.co/docs/transformers/tasks/summarization}}. All experiments were conducted on one node with 4 32GB V100 GPUs. The learning rate is set to 5e-5 for pre-trained models and is set to 1e-4 for Transformer. The pre-trained models are fine-tuned with 3 epochs, while the vanilla Transformer is trained with 20 epochs. For SAMSum, the maximum source length and target length is 512 and 90, and for DialogSum, EmailSum, QMSum, and TweetSumm, this setting is 512/150, 1,024/65, 2,048/200, and 1,024/120, respectively. The other hyper-parameters are set by default. 

\subsection{Details of Quality Assessment}
\label{appendix:human_evaluation}
\paragraph{Time Budget} We recruited three annotators to conduct the quality assessment for \textsc{Olds}. The total hits of judgment are 3000 (5 domains $\times$ 200 samples $\times$ 3 annotators). The annotating speed is 25 samples per hour and the workload is 120 hours (1000 / 25 * 3 = 120) in total.

\paragraph{Instructions} Each annotator was presented with a sample containing the dialogue, reference summary, candidate summary, gold oracles, candidate oracles, and the labeled omission utterances along with their corresponding omitted words. We instruct the annotators to make a binary choice whether the set of labeled omission utterances is {\em Accept} or {\em Reject}. Annotators should compare the candidate with the reference and find out omissions. Then, they should locate omissions in the original dialogue and record the corresponding utterances. Finally, they should compare the automatically labeled utterances with the recorded ones and make a judgment. The set of labeled omission utterances should be marked as {\em Reject} as long as it misses any critical utterance, or includes any redundant or uninformative utterance. Otherwise, it should be marked as {\em Accept}. To ensure that each choice is justified, we additionally asked annotators to perform corrections and renew the corresponding omitted words if the choice is {\em Reject}. Thus, we could verify why the labeled omission is marked as {\em Reject}. 

\subsection{Data Format}
\label{appendix:dataset_format}
To facilitate the community to explore the effect of possible elements on the omission problem, in the released version of \textsc{Olds}, we additionally provide some auxiliary information. Specifically, apart from the basic information of dialogue, reference summary, candidate summary, and omission labels, we further provide the intermediate information during labeling, including Gold Oracles, Candidate Oracles, omission words, and the source model and decoding strategy for each candidate summary, e.g., {\em `bart\_base, beam'}, which represents that the candidate is generated by BART$_{base}$ using beam search. A complete example is shown in Table~\ref{tb:full_example}.

\begin{table*}[t]
\fontsize{7.8pt}{8pt}\selectfont
\begin{center}
\setlength{\tabcolsep}{0.7mm}{
\begin{tabular}{cl|ccccccccccccccc}
\toprule[1pt]
\multirow{2}{*}{{\bf Model}}&\multirow{2}{*}{{\bf Metric}}&\multicolumn{3}{c}{\bf SAMSum} & \multicolumn{3}{c}{\bf DialogSum} & \multicolumn{3}{c}{\bf EmailSum} & \multicolumn{3}{c}{\bf QMSum} & \multicolumn{3}{c}{\bf TweetSumm} \\
& & Train & Dev. & Test & Train & Dev. & Test & Train & Dev. & Test & Train & Dev. & Test & Train & Dev. & Test \\
\midrule
\multirow{7}{*}{ALL} & OmitR. & 32.81 & 31.45& 31.93& 30.66 & 28.30&28.48 & 51.09 &41.10 & 42.52& 68.91& 74.71& 73.82& 62.96& 54.29& 55.91\\
& RG-1 & 43.47 & 48.47& 47.17&41.98 &45.22 &42.93 &28.89 & 34.22&33.61 &32.85 &30.90 &31.54 & 35.74& 45.10 & 42.32\\
& RG-2 & 17.97& 23.81& 22.21&16.21 &19.59 &16.81 &6.43 &10.33 &9.52 &10.48 & 9.88& 11.11& 13.77& 22.27 & 19.10\\
& RG-L & 33.84& 39.77& 38.60&32.63 &36.88 &33.84 & 21.89& 27.18& 26.55&21.31 & 21.15& 22.13& 27.16& 36.93 & 33.94\\
& BLEU & 10.66& 14.70& 13.60&16.05 &20.24 &16.94 &3.29 &5.69 & 5.50&7.22 & 6.15& 7.63& 9.98& 15.14 & 13.67\\
& BS$_\mathrm{B}$ & 90.56& 91.60& 91.45&91.02 &91.84 &91.41 &88.06 &89.20 &89.07 &86.33 & 86.30& 86.60& 86.96& 89.32 & 88.82\\
& BS$_\mathrm{L}$ & 90.51& 91.64& 91.48&90.19 &91.15 &90.74 &87.54 &88.70 & 88.61&85.71 & 85.87& 86.21& 86.35& 88.66 & 88.16\\
& BLEURT &48.32 &52.39 &52.17 &48.12 &51.06 &50.04 &47.47 &49.31 &49.06 &39.00 &39.53 &40.70 &45.73 &52.20 &50.93 \\
\midrule
\multirow{7}{*}{BART$_\mathrm{L}$} & OmitR. &21.60 & 21.48& 22.75&24.75 &22.38 &23.28 &33.65 &30.52 & 32.86&60.24 &61.40 &61.42 &42.30 &42.32 & 45.24\\
& RG-1 &50.62 & 54.38&52.57 &46.08 &49.55 &46.91 &33.16 &37.86 &36.98 &38.25 &35.94 &37.35 &47.25 &52.58 & 48.53\\
& RG-2 &25.29 & 29.91&28.00 &21.18 &25.11 &21.32 & 9.13&13.0 & 11.60&13.74 &12.74 &14.85 &22.19 & 28.72& 23.90\\
& RG-L &39.98 & 45.10&43.78 &36.50 &41.94 &37.97 & 24.72&30.04 &29.06 &23.83 &23.84 &25.58 &36.47 &43.89 & 39.42\\
& BLEU &14.34 & 19.55&17.88 &17.85 &23.23 &18.83 &4.02 &6.84 &6.28 &9.44 &7.94 &11.19 &15.39 &20.11 & 17.68\\
& BS$_\mathrm{B}$ &91.86 & 92.63&92.42 &91.41 &92.47 &92.05 &88.07 &89.86 &89.64 &87.52 &87.75 &88.13  &89.48 &90.67 & 89.92\\
& BS$_\mathrm{L}$ &91.89 & 92.71&92.50 &90.79 &91.96 &91.54 &87.61 &89.37 &89.18 &87.03 & 87.31&87.77 &88.82 &90.07 & 89.29\\
& BLEURT &56.54 &58.83 &57.95 &53.99 &56.38 &55.17 &49.14 &50.43 &50.08 &42.06 &41.58 &43.09 &54.15 &57.65 &55.30 \\
\midrule
\multirow{7}{*}{BART$_\mathrm{B}$} & OmitR. &27.21 &28.67 &30.56 &27.95 &26.57 &27.47 &37.01 &35.97 &37.49 & 62.74 &69.64 &69.46 & 41.05 & 45.48 & 46.37\\
& RG-1 &48.20 &51.12 &49.20 &44.14 &46.76 &44.33 &33.77 &36.63 &36.16 & 37.21 &33.87 &34.93 & 47.26 & 49.99 & 46.97\\
& RG-2 &22.70 &27.04 &24.41 &18.71 &21.95 &18.47 &9.34& 12.01&11.13 & 13.71 &11.39 &13.09 & 22.13 & 26.34 & 22.80\\
& RG-L &38.58 &42.73 &40.73 &34.99 &38.89 &35.11 &26.23 &29.20 &28.95 & 24.44 &22.79 &24.24 & 36.21 & 41.61 & 38.39\\
& BLEU &14.32 &16.17 &14.35 &18.15 &20.66 &17.67 &4.99 & 6.85&6.16 & 9.90 &5.71 &7.89 & 16.02 & 17.60 & 16.34\\
& BS$_\mathrm{B}$ &91.42 &92.21 &91.92 &91.63 &92.21 &91.67 &89.17 &89.66 &89.58 & 87.62 &87.55 &87.98 & 88.83 & 90.38 & 89.71\\
& BS$_\mathrm{L}$ &91.47 &92.26 &91.97 &90.99 &91.60 &91.15 &88.73 &89.22 &89.18 & 87.06 &87.09 &87.55 & 88.10 & 89.74 & 89.07\\
& BLEURT &53.87 &55.55 &54.80 &52.06 &53.59 &52.28 &48.80 &50.11 &49.67 &41.69 &39.85 &41.37 &53.31 &56.18 &54.31 \\
\midrule
\multirow{7}{*}{T5$_\mathrm{B}$} & OmitR. &21.95 &25.52 &26.82 &26.46 &25.90 &24.80 &28.67 &30.08 &32.39 &63.77 &68.14 &68.09 &43.89 &44.14 &45.51 \\
& RG-1 &47.18 &50.94 &49.29 &44.68 &46.21 &44.74 &33.03 &36.17 &35.36 &31.67 &33.05 &33.79 &43.16 &48.83 &44.38 \\
& RG-2 &21.70 &26.41 &24.17 &19.36 &20.48 &18.23 &9.58 &11.77 &10.83 &10.58 &11.40 &13.10 &20.60 &26.21 & 21.10\\
& RG-L &36.98 &42.27 &40.49 &35.26 &37.49 &35.26 &25.65 &28.86 &28.34 &21.58 &23.11 &24.29 &34.10 &41.11 & 36.08\\
& BLEU &12.00 &16.77 &15.40 &18.55 &20.59 &17.58 &4.23 &5.75 &5.75 &7.31 &7.49 &10.11 &14.56 &18.50 & 15.47\\
& BS$_\mathrm{B}$ &91.22 &92.00 &91.79 &91.68 &92.03 &91.68 &88.72 &89.44 &89.30 &86.16 &86.97 &87.26 &88.51 &90.32 & 89.32\\
& BS$_\mathrm{L}$ &91.22 &92.06 &91.83 &90.88 &91.38 &91.03 &88.23 &88.89 &88.82 &85.80 &86.51 &86.91 &87.91 &89.71 & 88.77\\
& BLEURT &53.37 &55.24 &54.68 &51.87 &52.63 &52.06 &49.70 &50.17 &50.05 &41.17 &40.64 &42.06 &50.99 &55.61 &52.91 \\
\midrule
\multirow{7}{*}{T5$_\mathrm{S}$} & OmitR. &33.16 &35.84 &34.81 &30.34 &31.46 &30.58 &35.91 &38.30 &39.59 &66.26 &76.05 &73.71 &53.21 &53.31 &53.63 \\
& RG-1 &41.03 &44.17 &43.53 &39.46 &40.26 &39.09 &29.71 &33.96 &32.39 &29.94 &28.07 &29.45 &30.69 &34.78 &32.62 \\
& RG-2 &16.76 &20.60 &19.37 &14.83 &15.17 &13.97 &7.35 &9.86 &9.08 &9.28 &8.19 &9.59 &11.93 &16.00 & 13.61\\
& RG-L &32.29 &36.30 &35.48 &31.00 &32.23 &30.67 &22.84 &27.09 &25.80 &20.14 &19.13 &20.37 &23.40 &28.02 & 25.29\\
& BLEU &9.40 &12.47 &11.55 &14.27 &15.79 &13.88 &3.37 &4.99 &5.24 &5.99 &5.76 &6.80 &8.35 &11.59 & 10.15\\
& BS$_\mathrm{B}$ &90.21 &90.90 &90.81 &90.74 &91.04 &90.78 &87.94 &88.81 &88.74 &85.33 &85.29 &85.74 &85.42 &86.67 & 86.58\\
& BS$_\mathrm{L}$ &90.24 &90.96 &90.87 &89.95 &90.30 &90.05 &87.42 &88.28 &88.22 &84.79 &84.82 &85.22 &85.15 &86.28 & 86.19\\
& BLEURT &47.66 &48.87 &49.25 &48.09 &48.34 &48.12 &47.86 &48.94 &48.09 &36.92 &36.94 &37.03 &42.49 &44.10 &44.49 \\
\midrule
\multirow{7}{*}{Transformer} & OmitR. &47.02 &48.43 &48.30 &39.91 &36.96 &38.89 &76.55 &76.29 &74.91 &87.76 &99.19 &94.72 &94.66 &95.46 &95.16 \\
& RG-1 &37.70 &39.14 &37.94 &39.09 &40.13 &36.76 &24.44 &24.79 &25.68 &29.52 &29.49 &29.11 &34.84 &35.44 & 36.03\\
& RG-2 &11.21 &12.53 &11.64 &10.75 &11.57 &9.07 &3.18 &3.25 &3.51 &7.03 &7.12 &6.98 &9.90 &10.45 & 11.39\\
& RG-L &28.31 &29.86 &29.11 &28.77 &30.41 &27.21 &18.14 &18.54 &18.74 &18.56 &18.92 &18.78 &25.67 &26.43 & 27.30\\
& BLEU &5.44 &5.87 &5.86 &12.14 &13.77 &10.92 &1.15 &1.27 &1.36 &2.71 &2.39 &2.36 &4.35 &4.21 & 5.03\\
& BS$_\mathrm{B}$ &89.46 &89.77 &89.67 &90.28 &90.76 &90.22 &87.64 &87.89 &87.88 &85.83 &85.93 &86.00 &87.57 &87.87 & 88.01\\
& BS$_\mathrm{L}$ &89.25 &89.66 &89.54 &89.16 &89.77 &89.23 &87.10 &87.31 &87.34 &84.71 &84.95 &85.08 &86.51 &86.84 & 86.88\\
& BLEURT &42.62 &43.33 &43.51 &42.13 &43.81 &42.03 &45.84 &45.59 &46.19 &37.96 &35.82 &36.52 &45.02 &46.15 &46.27 \\
\midrule
\multirow{7}{*}{Pegasus$_\mathrm{L}$} & OmitR. &- &28.75 &28.33 &- &26.56 &25.86 &- &35.46 &37.86 &- &73.85 &75.53 &- &45.04 & 49.51\\
& RG-1 &- &51.09 &50.42 &- &48.32 &45.80 &- &35.90 &35.19 &- &24.92 &24.63 &- &49.01 & 45.34\\
& RG-2 &- &26.44 &25.66 &- &23.27 &19.74 &- &12.04 &10.97 &- &8.41 &9.05 &- &25.79 & 21.91\\
& RG-L &- &42.44 &41.96 &- &40.32 &36.93 &- &29.33 &28.41 &- &19.09 &19.52 &- &40.35 & 37.07\\
& BLEU &- &16.24 &15.44 &- &23.84 &20.04 &- &6.85 &6.33 &- &3.90 &4.49 &- &17.68 & 16.41\\
& BS$_\mathrm{B}$ &- &92.10 &92.10 &- &92.51 &92.04 &- &89.52 &89.29 &- &84.31 &84.46 &- &89.98 & 89.40\\
& BS$_\mathrm{L}$ &- &92.21 &92.16 &- &91.90 &91.45 &- &89.13 &88.94 &- &84.55 &84.76 &- &89.35 & 88.78\\
& BLEURT &- &55.92 &56.16 &- &54.52 &53.49 &- &50.62 &50.31 &- &44.18 &45.88 &- &55.27 & 53.73\\
\bottomrule[1pt]
\end{tabular}}
\end{center}
\caption{\label{tb:dataset_results}{Detailed results of candidate summaries in the \textsc{Olds} dataset. OmitR. denotes {\em Omission Rate} in Eqation~\ref{eq:omission_rate}. RG stands for the ROUGE score. BS$_\mathrm{B}$ and BS$_\mathrm{L}$ denotes BERTScore using {\em RoBERTa-base} and {\em Roberta-large} as backbone models. For BLEURT, we use {\em BLEURT-20}. $\mathrm{L,B,S}$ in the subscript of model names stand for {\em large}, {\em base}, and {\em small} model sizes.}}
\end{table*}
\subsection{More Results of Candidate Summaries}
\label{appendix:dataset_results}
Table~\ref{tb:dataset_results} shows the evaluation results of candidate summaries in \textsc{Olds} assessed by various reference-based metrics. Here, we employ n-gram based metrics ROUGE~\cite{lin2004rouge} and BLEU~\cite{papineni2002bleu}, embedding-based metric BERTScore~\cite{zhang2019bertscore}, and learning-based metric BLEURT~\cite{sellam2020bleurt} to evaluate the candidate summaries.

\section{Omission Detection Models}
\subsection{Implementation Details}
\label{appendix:detection-setting}
We use BERT$_\mathrm{base}$ and RoBERTa$_\mathrm{base}$ as the backbone pre-trained encoder for the three frameworks. All the experiments were conducted on one node with a single A100 80GB GPU. For all three frameworks, the learning rate is set to 5e-5 and the training epoch is set to 5. The batch size was set to 128 for pair-wise classification and was set to 16 for sequence labeling and pointer network. We saved checkpoints after each epoch. The best performing checkpoint on the validation set was evaluated on the test set to report the final results.

\paragraph{Pair-wise Classification} For the framework of pair-wise classification, we use the official code of classification with pre-trained language models~\footnote{\url{https://huggingface.co/docs/transformers/tasks/sequence_classification}}. The input format is {\em <s>} $c$ {\em </s>} $u_i$ {\em </s>}, where {\em <s>} and {\em </s>} are classification token and separation token, respectively. $c$ and $u_i$ represent the candidate and the $i$-th utterance in the dialogue. 

\paragraph{Sequence Labeling} We use the same implementation as the extractive summarization model proposed by Liu and Lapata~\shortcite{liu2019text}. The only difference is that we append the candidate summary in front of the dialogue, denoted as {\em <s>} $c$ {\em </s>} {\em <s>} $u_1$ {\em </s>} {\em <s>} $u_2$ {\em </s>} ... {\em <s>} $u_N$ {\em </s>}. The {\em <s>} token before the candidate summary is not involved in the calculation. For SAMSum, we truncate each input into a maximum length of 512, while for DialogSum, EmailSum, QMSum, and TweetSumm, this setting is 512, 1,024, 2,048, and 1,024.

\paragraph{Pointer Network} The autoregressive decoder of our pointer network is implemented by a Transformer decoder, which is proposed by Zou et al. \shortcite{zou2021topic} and was previously used for extractive summarization. Here, we also append the candidate summary in front of the dialogue, which has the same input format as in sequence labeling. The {\em <s>} token before the candidate summary is not involved in the calculation. We also set the same maximum length as in sequence labeling for input sequences in different domains.

\begin{table*}[t!]
\setuldepth{Berlin}
\fontsize{7.5pt}{9.0pt}\selectfont
\begin{center}
\setlength{\tabcolsep}{0.6mm}{
\begin{tabular}{l}
\toprule[1pt]
\bf Dialogue: \\
(0) @AzureSupport Hi guys we have signed up a trial for log analytics while we test setting up a custom log import. The issue I have already is \\
\quad \quad trying to add a custom log import configuration, nothing is being added to the list, but no error messages? I have tried a hundred times.\\
(1) @242694 Could you please post here: -link- and send us the link so we can have an engineer on that team assist \\
(2) @AzureSupport Done.\\
(3) @242694 Could you please send us the link to the created post so we can alert the team? Thanks! \\
(4) @AzureSupport -link- \\
(5) @242694 Thank you! We have alerted the team and they should respond to your post shortly. \\
(6) @AzureSupport Thanks! \\
(7) @AzureSupport Hey guys no word yet, got a client waiting please. \\
(8) @242694 We're sorry about that. We've reached out again and will make sure that they reply to the forum post ASAP.  \\
(9) @AzureSupport -emoji- \\
(10) @AzureSupport No word yet \\
(11) @242694 We're sorry about this. We'll reach out to the team to ask where they are on this. \\
(12) @242694 A forum engineer has replied to your post. Please have a look and reply to the thread if you need further assistance. \\
\midrule[0.3pt]
\bf Reference summary: \\
\quad The customer says that he has signed up a trail for log analytics but he is having an issue in adding a custom log import configuration. The agent \\
asks to post the information regarding the isuue via provided link and to send them the link and says A forum engineer has replied to the post, \\
then asks to reply to the thread for further assistance. \\
\midrule[0.3pt]
\bf Candidate summaries: \\
\ [1] {\bf Source: bart-large, Strategy: beam} Customer is complaining that he is trying to add a custom log import configuration, nothing is being \\
\quad \quad added to the list, but no error messages. Agent updated that they have alerted the team and they should respond to their post shortly. \\
\ [2] {\bf Source: bart-large, Strategy: sample} Customer is complaining that he is trying to add a custom log import configuration, nothing is being \\
\quad \quad added to the list, but no error messages. Agent updated that they have alerted the team and they should respond to their post shortly. \\
\ [3] {\bf Source: bart-base, Strategy: beam} The customer says that he has signed up a trial for log analytics while he test setting up a custom log \\
\quad \quad import. The agent asks to send the link to the created post so that they can alert the team. \\
\ [4] {\bf Source: bart-base, Strategy: sample} Customer is complaining that they have signed up a trial for log analytics while they test setting up a \\
\quad \quad custom log import. Agent updated that they will reach out to the team to ask where they are on this issue. \\
\ [5] {\bf Source: t5-base, Strategy: beam} Customer is complaining that they have signed up a trial for log analytics while they test setting up a \\
\quad \quad custom log import. Agent updates that they have reached out again and will make sure that they reply to the forum post ASAP. \\
\ [6] {\bf Source: t5-base, Strategy: sample} Customer is complaining that they have signed up a trial for log analytics while they test setting up a \\
\quad \quad custom log import. Agent updates that they have reached out again and will make sure that they reply to the forum post ASAP. \\
\ [7] {\bf Source: t5-small, Strategy: beam} AzureSupport Hi guys we have signed up a trial for log analytics while we test setting up a custom log \\
\quad \quad import. The issue I have already is trying to add a custom log import configuration, nothing is being added to the list, but no error messages? \\
\quad \quad I have tried a hundred times. \\
\ [8] {\bf Source: t5-small, Strategy: sample} The issue I have already is trying to add a custom log import configuration, nothing is being added to the \\
\quad \quad list, but no error messages? I have tried a hundred times. Could you please post here: -link- and send us the link to the created post so we can \\
\quad \quad alert the team?\\
\ [9] {\bf Source: transformer, Strategy: beam} The customer says that he is unable to find the product he can't work with his phone. The agent asks \\
\quad \quad whether the customer is using and asks whether he will be able to send the issue and asks to assist further on the issue. \\
\ [10] {\bf Source: transformer, Strategy: sample} Customer is complaining that he is unable to know about the delay of the product. Agent updates \\
\quad \quad that they are unable to reach out for further assistance and requests to DM the issue.\\
\ [11] {\bf Source: pegasus, Strategy: beam} The issue I have already is trying to add a custom log import configuration, nothing is being added to the \\
\quad \quad list, but no error messages. The issue I have already is trying to add a custom log import configuration, nothing is being added to the list, but \\
\quad \quad no error messages. \\
\ [12] {\bf Source: pegasus, Strategy: sample} Customer is complaining that they have signed up a trial for log analytics while they are testing setting \\
\quad \quad up a custom log import. Agent updates that they have alerted the team and they should respond to their post shortly and adds that they have \\
\quad \quad reached out again and will make sure that they reply to the forum post ASAP.\\
\midrule[0.3pt]
{\bf Gold Oracles:} (0) (1) (3) (8) (11) (12) \\
\midrule[0.3pt]
{\bf Candidate Oracles:} \\
\ [1]: (0) (1) (5) (8) \quad \quad\quad\quad\quad\quad[4]: (0) (1) (5) (8) (11) \quad \quad\quad\quad\quad [7]: (0) \quad \quad\quad\quad\quad\quad \quad\quad\quad\quad\quad\quad\quad\quad\ \ [10]: (0) (3) (8) (11) (12)\\
\ [2]: (0) (1) (5) (8) \quad \quad\quad\quad\quad\quad [5]: (0) (1) (5) (8) (11) \quad \quad\quad\quad\quad[8]: (0) (1) (3) \quad \quad\quad\quad\quad\quad\quad\quad\quad\quad\quad \ \ \  [11]: (0) (3) (7) (12)\\
\ [3]: (0) (1) (3) (8) (11) (12) \quad \quad \ [6]: (0) (1) (5) (8) (11) \quad \quad\quad\quad\quad[9]: (0) (1) (3) (4) (8) (11) (12) \quad \quad\quad\quad\quad [12]: (0) (1) (5) (8) (11) (12) \\
\midrule[0.3pt]
\bf Omission utterances (Labels): \\
\ [1]: (0) (1) (12)\quad \quad\quad\quad\quad\quad\quad\  [4]: (0) (1) (12) \quad\quad\quad\quad\quad\quad\quad\quad [7]: (1) (12) \quad\quad\quad\quad\quad\quad\quad\quad\quad\quad\quad\quad\ \ [10]: (0) (1) (12)\\
\ [2]: (0) (1) (12)\quad \quad\quad\quad\quad\quad\quad\ [5]: (0) (1) (12) \quad\quad\quad\quad\quad\quad\quad\quad [8]: (0) (12) \quad\quad\quad\quad\quad\quad\quad\quad\quad\quad\quad\quad\ \   [11]: (0) (1) (12)\\
\ [3]: (0) (12) \quad\quad\quad\quad\quad\quad\quad\quad\ \   [6]: (0) (1) (12) \quad\quad\quad\quad\quad\quad\quad\quad [9]: (0) (1) (12) \quad\quad\quad\quad\quad\quad\quad\quad\quad\quad\quad  [12]: (0) (1) (12)\\
\midrule[0.3pt]
\bf Omission Words: \\
\ [1]: (0) issue, analytics, signed (1) engineer, send, link (12) engineer, forum, replied, assistance, thread, reply \\
\ [2]: (0) issue, analytics, signed (1) engineer, send, link (12) engineer, forum, replied, assistance, thread, reply\\
\ [3]: (0) issue, configuration (12) engineer, forum, replied, assistance, thread, reply\\
\ [4]: (0) configuration (1) engineer, post, send, link (12) engineer, forum, replied, assistance, thread, post, reply\\
\ [5]: (0) issue, configuration (1) engineer, send, link (12) engineer, replied, assistance, thread\\
\ [6]: (0) issue, configuration (1) engineer, send, link (12) engineer, replied, assistance, thread\\
\ [7]: (1) engineer, post, send, link (12) engineer, forum, replied, assistance, thread, post, reply\\
\ [8]: (0) analytics, signed (12) engineer, forum, replied, assistance, thread, reply\\
\ [9]: (0) analytics, custom, import, signed, log, configuration (1) engineer, post, link (12) engineer, forum, replied, assistance, thread, post, reply\\
\ [10]: (0) analytics, custom, import, signed, log, configuration (1) engineer, post, send, link (12) engineer, forum, replied, thread, post, reply\\
\ [11]: (0) analytics, signed (1) engineer, post, send, link (12) engineer, forum, replied, assistance, thread, post, reply\\
\ [12]: (0) issue, configuration (1) engineer, send, link (12) engineer, replied, assistance, thread\\
\bottomrule[1pt]
\end{tabular}}
\end{center}
\caption{\label{tb:full_example} A complete example in the \textsc{Olds} dataset, which is sampled from the test set of TweetSumm domain. }
\end{table*}

\end{document}